\documentclass[letterpaper, 10 pt, conference]{ieeeconf}  

\IEEEoverridecommandlockouts                              
\usepackage{amsmath} 
\usepackage{xcolor}
\usepackage{algorithm}
\usepackage{algpseudocode}
\usepackage{amssymb}
\usepackage{amsfonts}

\allowdisplaybreaks

\usepackage{caption, subcaption}
\usepackage{cite}
\usepackage{graphicx}

\title{\LARGE \bf
Kalman Bayesian Transformer 
}

\author{Haoming Jing, Oren Wright, Jos\'{e} M. F. Moura, and Yorie Nakahira
\thanks{The authors are with the Department of Electrical and Computer Engineering, Carnegie Mellon University, Pittsburgh, Pennsylvania, United States.
        {\tt\small \{haomingj, omw, moura, ynakahir\}@andrew.cmu.edu}}
\thanks{This work is sponsored in part by the PRESTO Grant Number JPMJPR2136 from the Japan Science and Technology Agency, in part by the National Science Foundation under Grants No. 2442948 and CCF-2327905, and in part by the Department of the Navy, Office of Naval Research, under award number N00014-23-1-2252. The views expressed are those of the authors and do not reflect the official policy or position of the US Navy, Department of Defense, or the US Government.}
}

\begin{document}

\maketitle
\thispagestyle{empty}
\pagestyle{empty}

\begin{abstract}
Sequential fine-tuning of transformers is useful when new data arrive sequentially, especially with shifting distributions. Unlike batch learning, sequential learning demands that training be stabilized despite a small amount of data by balancing new information and previously learned knowledge in the pre-trained models. This challenge is further complicated when training is to be completed in latency-critical environments and learning must additionally quantify and be mediated by uncertainty. 
Motivated by these challenges, we propose a novel method that frames sequential fine-tuning as a posterior inference problem within a Bayesian framework. Our approach integrates closed-form moment propagation of random variables, Kalman Bayesian Neural Networks, and Taylor approximations of the moments of softmax functions. By explicitly accounting for pre-trained models as priors and adaptively balancing them against new information based on quantified uncertainty, our method achieves robust and data-efficient sequential learning. The effectiveness of our method is demonstrated through numerical simulations involving sequential adaptation of a decision transformer to tasks characterized by distribution shifts and limited memory resources.  
\end{abstract}

\section{Introduction}

Efficient fine-tuning of transformer models has become increasingly important in machine learning applications. While pre-trained transformer models demonstrate remarkable performance across various tasks, they often experience degradation when deployed on data distributions that differ from their training sets. In these scenarios, full retraining can be prohibitively expensive, highlighting the need for lightweight, parameter-efficient fine-tuning methods~\cite{han2024parameter}. However, existing approaches typically require significant computational resources. 

Furthermore, data may arrive sequentially, and the onboard hardware may have limited memory to store all past data. In such scenarios, it is desirable to perform fine-tuning sequentially. Unlike batch learning, sequential learning requires the model to incorporate new information while preserving knowledge from past data. This balance is nontrivial to realize when only small amounts of data are available at each training step, amplifying the instability of training and the risk of catastrophic forgetting.

These challenges are further compounded by the need for quantifying uncertainty. Overconfident predictions can lead to catastrophic outcomes, particularly when data is limited or the system is intrinsically noisy. Incorporating uncertainty estimation (e.g., taking a Bayesian approach) also adds to the computational burden, making these methods even less tractable.

Bayesian approaches for transformer fine-tuning have emerged as promising alternatives that can explicitly incorporate prior knowledge and uncertainty~\cite{NEURIPS2023_cde2dc73, DBLP:journals/corr/abs-2104-08320, yang2024bayesianlowrankadaptationlarge}. However, conventional techniques for Bayesian deep learning---such as variational inference~\cite{graves2011variational, hoffman2013stochastic} and Markov Chain Monte Carlo (MCMC) sampling~\cite{metropolis1953equation, geman1984stochastic, neal1992bayesian, nemeth2021stochastic}---often involve computationally intensive iterative optimization with extensive sampling.

In this paper, we formulate fine-tuning as a posterior inference problem within a Bayesian framework and introduce the \textit{Kalman Bayesian Transformer}. Central to our method is a novel integration of closed-form characterization of neural network moments, Kalman Bayesian Neural Networks~\cite{wagner_kbnn_2023}, with a Taylor approximation of the distribution of a softmax function. Our method has the following advantages. 
\begin{itemize}
    \item \textbf{Sequential learning}: Our method explicitly incorporates previously trained parameters as a prior and balances the prior with new samples based on the size of uncertainty. This enables stable sequential training, which avoids the latency of waiting for future data and reduces onboard memory requirements (Figure~\ref{fig:success_rate}).
    \item \textbf{Computational efficiency}: Unlike traditional Bayesian methods that rely on expensive iterative sampling-based evaluations, our method evaluates closed-form formulas in a single pass, which significantly reduces the computational requirements (Figure~\ref{fig:computation_time}).
    \item \textbf{Explicit uncertainty quantification}: Our method quantifies the uncertainties in the prediction, which informs prediction confidence and enhances robustness against noisy data (Section~\ref{sec:fp}).
\end{itemize}

\section{Related work}

\textbf{Training transformers using Bayesian approaches.} Various Bayesian approaches have been developed for transformers. Some methods focus on learning all parameters with quantified uncertainties for training, while others explore parameter-efficient techniques for fine-tuning.
For example, Bayesian Transformer introduces a full Bayesian learning framework for transformers used in language modeling, which quantifies uncertainty and works with limited data~\cite{Bayesian-Transformer}. BayesFormer uses a dropout-based method for transformers that approximates variational inference~\cite{sankararaman2022bayesformertransformeruncertaintyestimation}. 
On the other hand, parameter-efficient techniques are studied, which aim to improve retraining efficiency by modifying only a subset of the transformer's parameters. For instance, BayesTune~\cite{NEURIPS2023_cde2dc73} employs a Laplace prior on each parameter to identify which parameters to update based on posterior estimates, and uses MCMC methods to draw samples from the posterior. 
BALM (Bayesian Active Learning with pre-trained language models) uses Monte Carlo dropout to approximate uncertainty, which is used in active learning to select data to annotate~\cite{DBLP:journals/corr/abs-2104-08320}. 
Bayesian low-rank adaptation uses Laplace approximation to estimate the posterior over adaptation parameters, demonstrating reduced overconfidence in models fine-tuned on small datasets~\cite{yang2024bayesianlowrankadaptationlarge}.
These approaches span a range of Bayesian techniques, ranging from variational inference, to MCMC, Monte Carlo dropout, and Laplace approximation. However, the existing literature for transformer (re)training has not explored the techniques from Kalman estimators.    

\textbf{Moment propagation techniques.} 
Various techniques have been proposed that propagate moments through nonlinear transformations in closed form~\cite{frey_variational_1999, boyen1998adf, minka2001}. A deterministic version of stochastic variational inference, proposed by~\cite{wu2019} and expanded on by~\cite{wright2024analytic}, uses moment propagation to solve variational inference in closed form. These techniques can provide the theoretical foundation and computational tools for efficient (re)training and analysis techniques of neural networks. For example, they are used to train probabilistic~\cite{gast2018lightweight} and fully Bayesian neural networks~\cite{hernandez-lobato2015pbp, ghosh2016assumed}. In this paper, we use the moment propagation method~\cite{wright2024analytic} to fine-tune transformers.  

\textbf{Bayesian neural networks.}
Bayesian deep learning, first proposed by~\cite{mackay1992}, can represent model uncertainty and allow for models that are robust and data-efficient. In contrast to standard neural networks, which find a single configuration of parameters $\mathcal{W}$ based on training data $\mathcal{D}$, Bayesian neural networks model the full posterior probability distribution $p(\mathcal{W} | \mathcal{D})$. This allows for the representation of predictive probability
\begin{equation*}
    p(\mathbf{y} | \mathbf{x}, \mathcal{D}) = \int_{\mathcal{W}} p(\mathbf{y} | \mathbf{x}, \mathcal{W}) p(\mathcal{W} | \mathcal{D}) d\mathcal{W}.
\end{equation*}
This predictive probability distribution, or Bayesian model average, accounts for epistemic uncertainty (analogous to measurement noise in control), i.e. the uncertainty over which parameter setting is correct given the observed data. This can be reduced with additional training data, in contrast to aleatoric uncertainty, which is inherent in the underlying data process (analogous to system noise in control) and is irreducible~\cite{kendall2017uncertainties, hullermeier2021aleatoric}. However, exact Bayesian inference is too computationally expensive for modern neural networks, and extensive efforts have been made to develop approximate Bayesian techniques with improved computation~\cite{izmailov2021bayesian}. Markov chain Monte Carlo sampling has been widely studied, and approximates the posterior by sampling from a Markov process~\cite{metropolis1953equation, geman1984stochastic, neal1992bayesian, nemeth2021stochastic}. Variational inference~\cite{graves2011variational, hoffman2013stochastic} and Laplace approximation~\cite{mackay1992, kristiadi2021learnable} use differential information to fit a Gaussian distribution to the posterior. Kalman smoothing is applied to train Bayesian neural networks using closed-form formulas~\cite{wagner_kbnn_2023}. Techniques like dropout~\cite{gal2016dropout} and SWAG~\cite{maddox2019swag} can be used to cheaply approximate the variational distribution. We integrate \cite{wagner_kbnn_2023} with moment propagation~\cite{wright2024analytic} and approximation techniques of distributions for a different problem, sequentially fine-tuning transformers.

\section{Problem Statement}
We consider the problem of fine-tuning a transformer model~\cite{vaswani2017attention} in a supervised learning setting. Below, we introduce the notation, the transformer model, and the Bayesian fine-tuning problem. 

\textbf{Notation.}
We denote scalars, vectors, and matrices by $x$, $\mathbf{x}$, and $\mathbf{X}$, respectively. Depending on the context, $\mathbf{X}$ may also denote a set of matrices and/or vectors. We use $\boldsymbol{0}_{d_1\times d_2}$ to denote a $d_1$ by $d_2$ matrix with all entries $0$. We use $\boldsymbol{1}_{d_1\times d_2}$ to denote a $d_1$ by $d_2$ matrix with all entries $1$. We use $\mathbf{I}_{d_1}$ to denote a $d_1$ by $d_1$ identity matrix. We may omit the subscript when the dimension is obvious. We use $\boldsymbol{\mu}_{\mathbf{v}}$ to denote the mean of a random vector $\mathbf{v}$, $\boldsymbol{\Sigma}_{\mathbf{v},\mathbf{v}}$ to denote the covariance of a random vector $\mathbf{v}$, and $\boldsymbol{\Sigma}_{\mathbf{v}_1,\mathbf{v}_2}$ to denote the cross-covariance between two random vectors $\mathbf{v}_1$ and $\mathbf{v}_2$. The variables $i$, $j$, $k$ and $l$ are exclusively used as indices, such as the lo op counter in the pseudocode. We use $\mathbf{A}[i,j]$ to denote the real number at the $i$-th row and $j$-th column of a matrix $\mathbf{A}$, and use $\mathbf{a}[i]$ to denote the real number at the $i$-th entry of a vector $\mathbf{a}$. We use $\mathbf{A}[i:j,k:l]$ to denote the submatrix that is formed by the rows $i$ to $j$ (inclusive) and columns $k$ to $l$ (inclusive) of matrix $\mathbf{A}$. We use $\lceil\rceil$ and $\lfloor\rfloor$ to denote the ceiling and flooring functions on real numbers, \textit{i.e.,} $\lceil x\rceil=\min\{y\in\mathbb{Z}\mid y\geq x\}$ and $\lfloor x\rfloor=\max\{y\in\mathbb{Z}\mid y\leq x\}$.

\begin{figure}
    \centering
    \includegraphics[width=\linewidth]{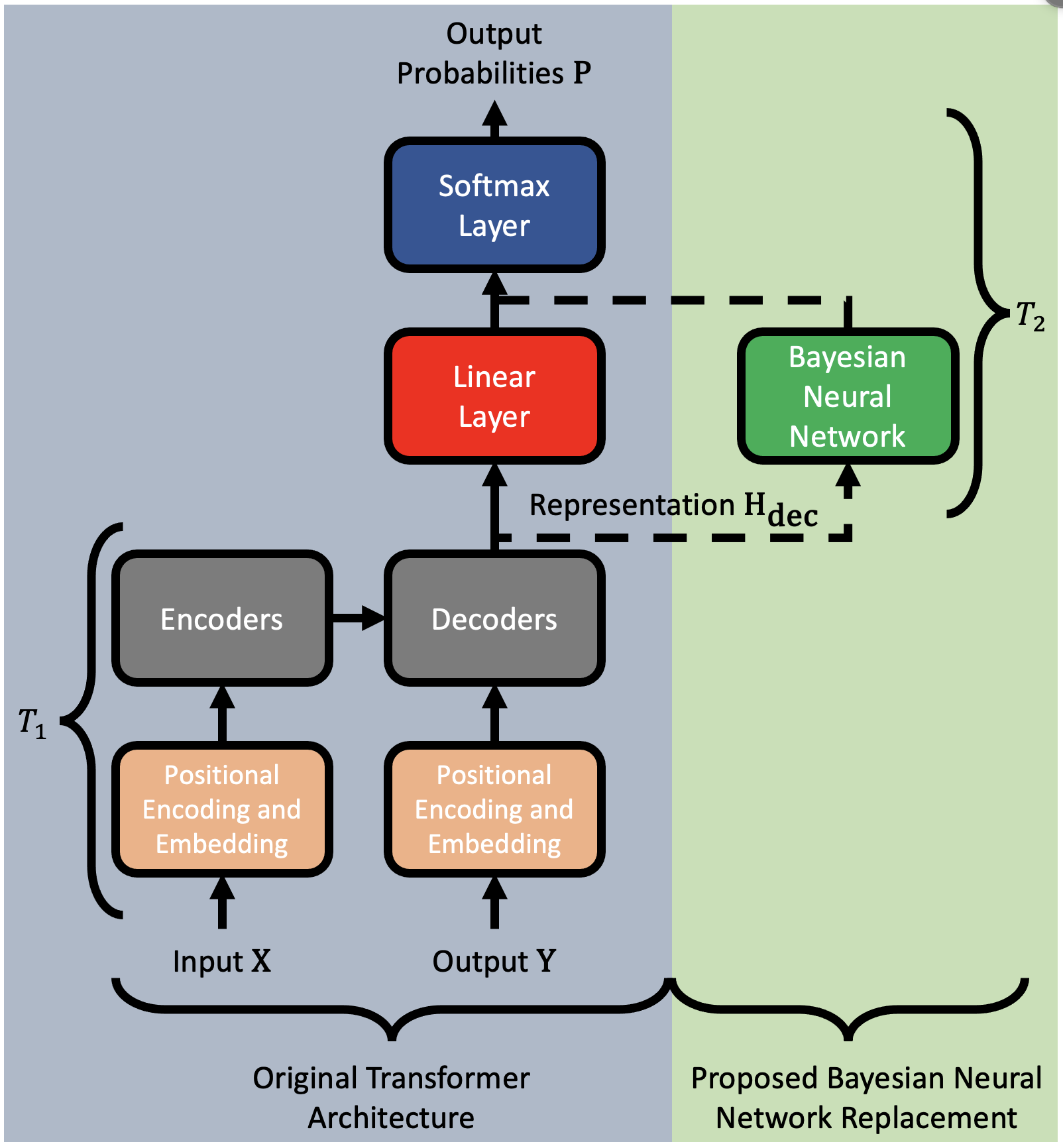}
    \caption{Proposed architecture. The original transformer architecture is shown on the left, and the Bayesian neural network that replaces the linear layer is shown on the right.}
    \label{fig:transformer_overview}
\end{figure}

\textbf{Transformer.}
 We consider a known and pre-trained transformer network which can be represented as a mapping $T:\mathbb{R}^{x\times d}\times\mathbb{R}^{y\times d}\rightarrow (0,1)^{y\times d_o}$, where $x$ is the length of the input sequence, $y$ is the length of the output sequence, $d$ is the embedding dimension, and $d_o$ is the output dimension. We separate the network into 2 parts, \textit{i.e.,} $T=T_2\circ T_1$, where $T_1:\mathbb{R}^{x\times d}\times\mathbb{R}^{y\times d}\rightarrow\mathbb{R}^{y\times d}$ is the mapping from the input sequence and target sequence to the representation $\mathbf{H}_{\text{dec}}=[\mathbf{h}^i_1,\mathbf{h}^i_2,\cdots,\mathbf{h}^i_{y}]^T$, and $T_2:\mathbb{R}^{y\times d}\rightarrow (0,1)^{y\times d_o}$ is the mapping from the representation $\mathbf{H}_{\text{dec}}$ to the output probability $\mathbf{P}$, as shown in Figure~\ref{fig:transformer_overview}. Since the transformer outputs a probability distribution, we define a one-hot mapping $f_{\text{one-hot}}:\mathbb{R}^d\rightarrow \{0,1\}^{d_{\text{world}}}$ that maps a vectorized token to a vector of $0$s and a $1$, where the $1$ appears in the entry corresponding to the token in the dictionary. 

\begin{figure}
    \centering
    \includegraphics[width=\linewidth]{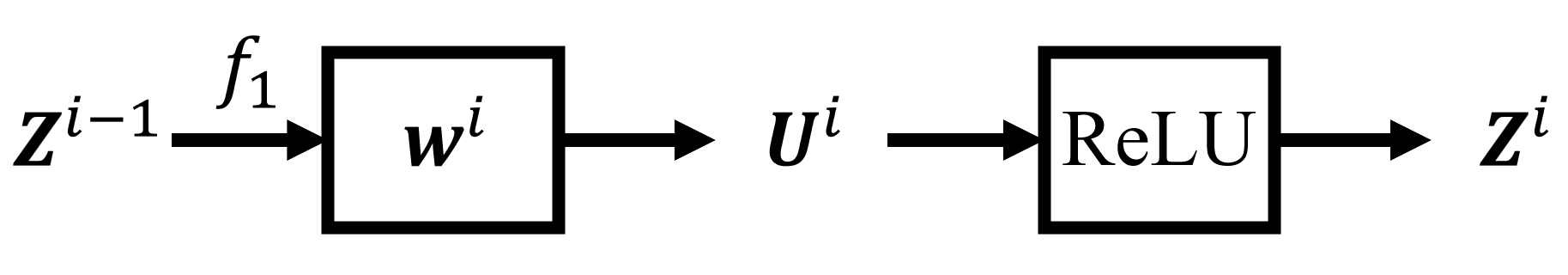}
    \caption{Illustration for one layer of the Bayesian neural network (\eqref{eq:u_def} and \eqref{eq:z_def}).}
    \label{fig:layers_illustration}
\end{figure}

\textbf{Bayesian fine-tuning.}
The training data available for fine-tuning are $\mathcal{D}=\{\{\mathbf{X}^1,\mathbf{Y}^1\},\{\mathbf{X}^2,\mathbf{Y}^2\},\cdots,\{\mathbf{X}^N,\mathbf{Y}^N\}\}$, where $N$ is the number of sequences available for fine-tuning. Regarding the $i$-th sequence, $\mathbf{X}_i\in\mathbb{R}^{x_i\times d}$ is the input sequence, and $\mathbf{Y}^i\in\mathbb{R}^{y_i\times d}$ is the output sequence. Here, $x_i$ is the length of the input sequence, $y_i$ is the length of the output sequence, and $d$ is the dimension of the embedding. We assume that all sets of sequences in the training data are independent and identically distributed. 
In the fine-tuning process, we freeze $T_1$ and replace the linear part in $T_2$ with a Bayesian neural network with $L$ layers. A ReLU activation layer is placed in between 2 linear layers. Let $\mathbf{U}^i=[\mathbf{u}^i_1,\mathbf{u}^i_2,\cdots,\mathbf{u}^i_{y}]^T\in\mathbb{R}^{y\times n_i},i\in\{1,2,\cdots,L\}$ be the intermediate representation after the $i$-th linear layer, and $\mathbf{Z}^i=[\mathbf{z}^i_1,\mathbf{z}^i_2,\cdots,\mathbf{z}^i_{y}]^T\in\mathbb{R}^{y\times n_i},i\in\{1,2,\cdots,L-1\}$ be the intermediate representation after the $i$-th ReLU activation layer, where $n_i$ is the number of neurons in the $i$-th layer. The inference process through the Bayesian neural network is given by
\begin{align}
    \mathbf{w}^i&=[(\mathbf{w}^i_1)^T,(\mathbf{w}^i_2)^T,\cdots,(\mathbf{w}^i_{n_i})^T]^T\sim p(\mathbf{w}^i)\\
    \mathbf{U}^i&=f_1(\mathbf{Z}^{i-1})[\mathbf{w}^i_1,\mathbf{w}^i_2,\cdots,\mathbf{w}^i_{n_i}]\label{eq:u_def}\\
    \mathbf{Z}^{i}&=\max (\boldsymbol{0},\mathbf{U}^i),\label{eq:z_def}
\end{align}
for all $i\in\{1,2,\cdots,L\}$, where the function $f_1$ appends a vector of $1$s to account for the bias term. An illustration of this structure is given in Figure~\ref{fig:layers_illustration}. The input $\mathbf{Z}^0$ to the network is $\mathbf{H}_{\text{dec}}$, and the output is given by
\begin{align}
    \mathbf{P}=[\mathbf{p}_1,\mathbf{p}_2,\cdots,\mathbf{p}_{y}]^T=\text{softmax}(\mathbf{U}^L).
\end{align}
The objective is to computationally efficiently find the posterior distribution of the weights
\begin{align}
\label{eq:posterior_def}
    p(\mathbf{W}_{\text{B}}|\mathcal{D})=\frac{p(\mathcal{D}|\mathbf{W}_{\text{B}})p(\mathbf{W}_{\text{B}})}{p(\mathcal{D})},
\end{align}
where $\mathbf{W}_{\text{B}}=\{\mathbf{w}^1,\mathbf{w}^2,\cdots,\mathbf{w}^L\}\in\mathcal{W}$ is the set of parameters in the Bayesian neural network~\cite{mackay1992,neal1992belief,wagner_kbnn_2023}. 

\section{Proposed Method}
Following~\cite{wagner_kbnn_2023}, we use a Bayesian smoothing algorithm to approximate the neural network's posterior distribution by propagating first- and second-order moments. Let $\mathcal{D}^k:=\{\{\mathbf{X}^1,\mathbf{Y}^1\},\{\mathbf{X}^2,\mathbf{Y}^2\},\cdots,\{\mathbf{X}^k,\mathbf{Y}^k\}\}$. We have the following factorization of \eqref{eq:posterior_def}:
\begin{align}
    p(\mathbf{W}_{\text{B}}|\mathcal{D}^k)=&\frac{p(\mathcal{D}^k|\mathbf{W}_{\text{B}})p(\mathbf{W}_{\text{B}})}{p(\mathcal{D}^k)}\\
    =&\frac{\prod^k_{i=1}p(\mathbf{X}^i,\mathbf{Y}^i|\mathbf{W}_{\text{B}})p(\mathbf{W}_{\text{B}})}{\prod^k_{i=1}p(\mathbf{X}^i,\mathbf{Y}^i)}\label{eq:iid_substitution}\\
    =&\frac{\prod^{k-1}_{i=1}p(\mathbf{X}^i,\mathbf{Y}^i|\mathbf{W}_{\text{B}})p(\mathbf{X}^k,\mathbf{Y}^k|\mathbf{W}_{\text{B}})p(\mathbf{W}_{\text{B}})}{\prod^k_{i=1}p(\mathbf{X}^i,\mathbf{Y}^i)}\\
    =&\frac{p(\mathcal{D}^{k-1}|\mathbf{W}_{\text{B}})p(\mathbf{X}^k,\mathbf{Y}^k|\mathbf{W}_{\text{B}})p(\mathbf{W}_{\text{B}})}{\prod^k_{i=1}p(\mathbf{X}^i,\mathbf{Y}^i)}\\
    =&\frac{\frac{p(\mathbf{W}_{\text{B}}|\mathcal{D}^{k-1})p(\mathcal{D}^{k-1})}{p(\mathbf{W}_{\text{B}})}p(\mathbf{X}^k,\mathbf{Y}^k|\mathbf{W}_{\text{B}})p(\mathbf{W}_{\text{B}})}{\prod^k_{i=1}p(\mathbf{X}^i,\mathbf{Y}^i)}\\
    =&\frac{p(\mathbf{W}_{\text{B}}|\mathcal{D}^{k-1})p(\mathbf{X}^k,\mathbf{Y}^k|\mathbf{W}_{\text{B}})\prod^{k-1}_{i=1}p(\mathbf{X}^i,\mathbf{Y}^i)}{\prod^k_{i=1}p(\mathbf{X}^i,\mathbf{Y}^i)}\\
    =&\frac{p(\mathbf{W}_{\text{B}}|\mathcal{D}^{k-1})p(\mathbf{X}^k,\mathbf{Y}^k|\mathbf{W}_{\text{B}})}{p(\mathbf{X}^k,\mathbf{Y}^k)},\label{eq:bayesian_factorization_final}
\end{align}
where \eqref{eq:iid_substitution} is due to the assumption that the training data are i.i.d. Therefore, from \eqref{eq:bayesian_factorization_final}, we have
\begin{align}
\label{eq:posterior_factorization}
    p(\mathbf{W}_{\text{B}}|\mathcal{D}^k)\propto p(\mathbf{W}_{\text{B}}|\mathcal{D}^{k-1})p(\mathbf{X}^k,\mathbf{Y}^k|\mathbf{W}_{\text{B}}),
\end{align}
which indicates that, given the distribution $p(\mathbf{W}_{\text{B}}|\mathcal{D}^{k-1})$, obtaining the distribution $p(\mathbf{W}_{\text{B}}|\mathcal{D}^k)$ only requires $\{\mathbf{X}^k,\mathbf{Y}^k\}$. Therefore, we can update the weight iteratively. Each update consists of 2 parts:
\begin{itemize}
    \item Forward Pass. The forward pass uses Bayesian filtering computes the distributions of the intermediate variables and output, \textit{i.e.,} $\{\mathbf{U}^i\}_{i\in\{1,2,\cdots,L\}}$, $\{\mathbf{Z}^i\}_{i\in\{1,2,\cdots,L\}}$ and $\mathbf{P}$, conditioned on $\mathbf{Z}^0=\mathbf{H}_{\text{dec}}$ and the weights $\mathbf{W}_{\text{B}}$.
    \item Backward Pass. The backward pass uses Kalman smoothing to update the distributions of the intermediate variables and weights in a backward manner, \textit{i.e.,} from layer $L$ to layer $1$, given the target output as new measurement.
\end{itemize}
For reasons of economy, we will omit bias terms in the remainder of this section without loss of generality, as a layer's input and weight matrix can be straightforwardly augmented to encompass the bias.

\subsection{Training Data Preprocessing}
Due to the autoregressive nature of the transformer, each pair of training data $\{\mathbf{X}^k,\mathbf{Y}^k\},\mathbf{X}^k\in\mathbb{R}^{x_k\times d},\mathbf{Y}^k\in\mathbb{R}^{y_k\times d}$ can be seen as generated by a sequential process of $y_k-1$ steps, where the $i$-th step takes input $\{\mathbf{X}^k,\mathbf{Y}^k[1:i,1:d]\}$, obtains output $\mathbf{P}$ from the transformer, and uses $\mathbf{P}[i:i,1:d_o]$ to determine the distribution of the $(i+1)$-th element in the output\footnote{When applicable, the first row in the output $\mathbf{Y}^k$ always corresponds to the start-of-sequence (SOS). Therefore, in the first step ($i=1$), the second element in the sequence is generated.}. If each step is considered a training instance, then training instances are not i.i.d. The recursive Bayesian update \eqref{eq:posterior_factorization} requires each update of $p(\mathbf{W}_{\text{B}}|\mathcal{D}^k)$, \textit{i.e.,} each forward pass and backward pass, to take into account the complete pair of training data $\{\mathbf{X}^k,\mathbf{Y}^k\}$. We therefore modify the data representation with Algorithm~\ref{alg:preprocessing}, which preprocesses the data such that the data within a training pair are converted to a single batched representation, independent from other batches, to be propagated in the forward pass and backward pass. Specifically, the input for each training instance is converted to the representation $\mathbf{H}_{\text{dec}}$ (line 4), and the output is converted to the one-hot representation (line 6). The procedure returns the batched representations corresponding to $\{\mathbf{X}^k,\mathbf{Y}^k\}$ (line 9). Each sequence of augmented data can be treated as i.i.d. and \eqref{eq:posterior_factorization} now holds.

\begin{algorithm}
\caption{Data Preprocessing}
\label{alg:preprocessing}
\begin{algorithmic}[1]
\Require trained transformer $T=T_2\circ T_1$ with weights
\Procedure{Preprocessing}{$\{\mathbf{X}^k,\mathbf{Y}^k\}$}
\State Initialize $\hat{\mathbf{H}}$ and $\hat{\mathbf{Y}}$ to be empty $0$ by $d$ matrices
\For{$i$ in $\{1,2,\cdots, y_k-1\}$}
\State $\hat{\mathbf{H}}\gets [\hat{\mathbf{H}}^T,T_1(\mathbf{X}^k,\mathbf{Y}^k[1:i,1:d])^T]^T$
\For{$j$ in $\{2,3,\cdots,i+1\}$}
\State $\hat{\mathbf{Y}}\gets [\hat{\mathbf{Y}}^T,f_{\text{one-hot}}(\mathbf{Y}^k[j:j,1:d]^T)]^T$
\EndFor
\EndFor
\State \textbf{return} $\hat{\mathbf{H}},\hat{\mathbf{Y}}$
\EndProcedure
\end{algorithmic}
\end{algorithm}

\subsection{Forward Pass}
\label{sec:fp}
Because the distribution of $\mathbf{U}^i$ and $\mathbf{Z}^i$ in layer $i$ only depends on the distribution of weights at layer $i$ and the output $\mathbf{Z}^{i-1}$ from the previous layer, we can propagate the first and second moment (mean and variance) one layer at a time. We define the vectorization
\begin{align}
    \mathbf{u}^i:=&[(\mathbf{u}^i_1)^T,(\mathbf{u}^i_2)^T,\cdots,(\mathbf{u}^i_{y})^T]^T\\
    \mathbf{z}^i:=&[(\mathbf{z}^i_1)^T,(\mathbf{z}^i_2)^T,\cdots,(\mathbf{z}^i_{y})^T]^T\\
    \mathbf{p}:=&[(\mathbf{p}_1)^T,(\mathbf{p}_2)^T,\cdots,(\mathbf{p}_{y})^T]^T
\end{align}
for all layers $i$. Let
\begin{align}
    \tilde{\mathbf{W}}^i=\begin{bmatrix}
        (\mathbf{w}^i_1)^T & \boldsymbol{0} & \cdots & \boldsymbol{0}\\
        \boldsymbol{0} & (\mathbf{w}^i_2)^T & \cdots & \boldsymbol{0}\\
        \vdots & \vdots & \ddots &\vdots\\
        \boldsymbol{0} & \boldsymbol{0} & \cdots & (\mathbf{w}^i_{n_i})^T
    \end{bmatrix}.
\end{align}
The linear transformation can be written as
\begin{align}
\label{eq:linear_u_to_w}
    \mathbf{u}^i=\text{diag}(\underbrace{\tilde{\mathbf{W}}^i,\tilde{\mathbf{W}}^i,\cdots,\tilde{\mathbf{W}}^i}_{y})\mathbf{z}^{i-1}:=\Bar{\mathbf{W}}^i\mathbf{z}^{i-1}.
\end{align}
Assuming $\mathbf{w}^i$ and $\mathbf{z}^{i-1}$ are independent, the forward pass through the linear layer $i$ is given by
\begin{align}
    \boldsymbol{\mu}_{\mathbf{u}^i}=&\Bar{\mathbf{M}}^i\boldsymbol{\mu}_{\mathbf{z}^{i-1}},\label{eq:mu_u_forward}
\end{align}
where $\Bar{\mathbf{M}}^i$ is the mean of $\Bar{\mathbf{W}}^i$. Given
\begin{align}
    \mathbf{u}^i[j]=(\mathbf{w}^i_{j-n_i\lfloor\frac{j}{n_i}\rfloor})^T\mathbf{z}^{i-1}_{\lceil\frac{j}{n_i}\rceil},
\end{align}
the covariance $\boldsymbol{\Sigma}_{\mathbf{u}^i,\mathbf{u}^i}$ can be easily calculated with 
\begin{align}
\label{eq:sigma_u_forward}
    \boldsymbol{\Sigma}_{\mathbf{u}^i,\mathbf{u}^i}[j,k]=\mathbb{E}[\mathbf{u}^i[j]\mathbf{u}^i[k]]-\boldsymbol{\mu}_{\mathbf{u}^i}[j]\boldsymbol{\mu}_{\mathbf{u}^i}[k]
\end{align}
for all $j,k\in\{1,2,\cdots,n_i\}$ using $\boldsymbol{\mu}_{\mathbf{z}^{i-1}}$, $\boldsymbol{\mu}_{\mathbf{w}^i}$, $\boldsymbol{\Sigma}_{\mathbf{z}^{i-1},\mathbf{z}^{i-1}}$, and $\boldsymbol{\Sigma}_{\mathbf{w}^i,\mathbf{w}^i}$. From~\cite{wagner_kbnn_2023} and~\cite[Theorem 1]{wright2024analytic}, the forward pass through the ReLU layer $i$ is given by
\begin{align}
    \boldsymbol{\mu}_{\mathbf{z}^i}[j]=&\boldsymbol{\mu}_{\mathbf{u}^i}[j]\Phi\left(\frac{\boldsymbol{\mu}_{\mathbf{u}^i}[j]}{\sqrt{\boldsymbol{\Sigma}_{\mathbf{u}^i,\mathbf{u}^i}[j,j]}}\right)\nonumber\\
    &+\sqrt{\boldsymbol{\Sigma}_{\mathbf{u}^i,\mathbf{u}^i}[j,j]}\phi\left(\frac{\boldsymbol{\mu}_{\mathbf{u}^i}[j]}{\sqrt{\boldsymbol{\Sigma}_{\mathbf{u}^i,\mathbf{u}^i}[j,j]}}\right)\label{eq:mu_z_forward}
\end{align}
\begin{align}
    &\boldsymbol{\Sigma}_{\mathbf{z}^i,\mathbf{z}^i}[j,k]=\nonumber\\
    &\Phi\left(\frac{\boldsymbol{\mu}_{\mathbf{u}^i}[j]}{\sqrt{\boldsymbol{\Sigma}_{\mathbf{u}^i,\mathbf{u}^i}[j,j]}}\right)\boldsymbol{\Sigma}_{\mathbf{u}^i,\mathbf{u}^i}[j,k]\Phi\left(\frac{\boldsymbol{\mu}_{\mathbf{u}^i}[k]}{\sqrt{\boldsymbol{\Sigma}_{\mathbf{u}^i,\mathbf{u}^i}[k,k]}}\right)\nonumber\\
    &+\frac{1}{2\sqrt{\boldsymbol{\Sigma}_{\mathbf{u}^i,\mathbf{u}^i}[j,j]\boldsymbol{\Sigma}_{\mathbf{u}^i,\mathbf{u}^i}[k,k]}}\phi\left(\frac{\boldsymbol{\mu}_{\mathbf{u}^i}[j]}{\sqrt{\boldsymbol{\Sigma}_{\mathbf{u}^i,\mathbf{u}^i}[j,j]}}\right)\nonumber\\
    &\boldsymbol{\Sigma}_{\mathbf{u}^i,\mathbf{u}^i}[j,k]\phi\left(\frac{\boldsymbol{\mu}_{\mathbf{u}^i}[k]}{\sqrt{\boldsymbol{\Sigma}_{\mathbf{u}^i,\mathbf{u}^i}[k,k]}}\right)\label{eq:sigma_z_forward}
\end{align}
for all $j,k\in\{1,2,\cdots,n_i\}$, where $\Phi$ and $\phi$ are the Gaussian cumulative density function and Gaussian probability density function, respectively. For brevity \eqref{eq:sigma_z_forward} shows only a second-order expansion, but the covariance can be computed to arbitrary precision~\cite{wright2024analytic}. Because the softmax is a vector operation for which mean and covariance cannot be easily computed, we use a first-order Taylor approximation to compute the forward pass through the softmax layer. Let $f_s$ be the softmax function. When the variance is small, the first order Taylor expansion about $\boldsymbol{\mu}_{\mathbf{u}^L}$ is given by
\begin{align}
    f_s(\mathbf{u}^L)\approx f_s(\boldsymbol{\mu}_{\mathbf{u}^L})+\mathbf{J}(\boldsymbol{\mu}_{\mathbf{u}^L})(\mathbf{u}^L-\boldsymbol{\mu}_{\mathbf{u}^L}).
\end{align}
The approximated mean for $\mathbf{p}=f_s(\mathbf{u}^L)$ is given by
\begin{align}
    \boldsymbol{\mu}_{\mathbf{p}}=&\mathbb{E}[f_s(\mathbf{u}^L)]\\
    \approx &\mathbb{E}[f_s(\boldsymbol{\mu}_{\mathbf{u}^L})+\mathbf{J}(\boldsymbol{\mu}_{\mathbf{u}^L})(\mathbf{u}^L-\boldsymbol{\mu}_{\mathbf{u}^L})]\\
    =&f_s(\boldsymbol{\mu}_{\mathbf{u}^L})\label{eq:softmax_mu_forward}
\end{align}
since $f_s(\boldsymbol{\mu}_{\mathbf{u}^L})$ and $\mathbf{J}(\boldsymbol{\mu}_{\mathbf{u}^L})$ are not functions of $\mathbf{u}^L$, and $\mathbb{E}(\mathbf{u}^L)=\boldsymbol{\mu}_{\mathbf{u}^L}$. Here, $\mathbf{J}$ is the Jacobian of the softmax function\footnote{With slight abuse of terms, we refer here the function that individually applies softmax to each sub-vector $\mathbf{u}^L_j$ before vectorization.}. Similarly, we have
\begin{align}
    \boldsymbol{\Sigma}_{\mathbf{p},\mathbf{p}}=&\mathbb{E}[(f_s(\mathbf{u}^L)-\boldsymbol{\mu}_{\mathbf{p}})(f_s(\mathbf{u}^L)-\boldsymbol{\mu}_{\mathbf{p}})^T]\\
    \approx &\mathbb{E}[(f_s(\mathbf{u}^L)-f_s(\boldsymbol{\mu}_{\mathbf{u}^L}))(f_s(\mathbf{u}^L)-f_s(\boldsymbol{\mu}_{\mathbf{u}^L}))^T]\\
    \approx & \mathbb{E}[(\mathbf{J}(\boldsymbol{\mu}_{\mathbf{u}^L})(\mathbf{u}^L-\boldsymbol{\mu}_{\mathbf{u}^L}))(\mathbf{J}(\boldsymbol{\mu}_{\mathbf{u}^L})(\mathbf{u}^L-\boldsymbol{\mu}_{\mathbf{u}^L}))^T]\\
    =& \mathbf{J}(\boldsymbol{\mu}_{\mathbf{u}^L})\mathbb{E}[(\mathbf{u}^L-\boldsymbol{\mu}_{\mathbf{u}^L})(\mathbf{u}^L-\boldsymbol{\mu}_{\mathbf{u}^L})^T]\mathbf{J}(\boldsymbol{\mu}_{\mathbf{u}^L})^T\\
    =&\mathbf{J}(\boldsymbol{\mu}_{\mathbf{u}^L})\boldsymbol{\Sigma}_{\mathbf{u}^L,\mathbf{u}^L}\mathbf{J}(\boldsymbol{\mu}_{\mathbf{u}^L})^T\label{eq:softmax_sigma_forward}
\end{align}
and 
\begin{align}
    \boldsymbol{\Sigma}_{\mathbf{u}^L,\mathbf{p}}=&\boldsymbol{\Sigma}_{\mathbf{u}^L,\mathbf{u}^L}\mathbf{J}(\boldsymbol{\mu}_{\mathbf{u}^L})^T.\label{eq:softmax_cross_covariance}
\end{align}
We note that the accuracy of this approximation can be improved with the addition of higher order Taylor expansion terms.

\subsection{Backward Pass}
\subsubsection{Weight Initialization}
The iterative update of $p(\mathbf{W}_{\text{B}}|\mathcal{D}^k)$ starts with $k=0$. We would like $p(\mathbf{W}_{\text{B}}|\mathcal{D}^0)=p(\mathbf{W}_{\text{B}})$ to replicate the behavior of the linear layer of the original transformer. To achieve that, we design $p(\mathbf{W}_{\text{B}})$ such that $\mathbf{W}_{\text{B}}$ is almost deterministic ($\boldsymbol{\Sigma}_{\mathbf{w}^i,\mathbf{w}^i} = \epsilon \mathbf{I}$ for all $i\in\{1,2,\cdots,L\}$, where $\epsilon$ is a hyperparameter that prevents singularity in matrix inversion), and the input $\mathbf{z}^0$ to the network is preserved until the last layer, where the weight $\mathbf{W}_{\text{O}}$ of the linear layer in the transformer is applied. To preserve the input through the ReLU activation, we design the initial weight such that $\mathbf{u}^i,\forall i\in\{1,2,\cdots,L-1\}$ contains a positive copy and a negative copy of the input. In this way, both the positive and negative parts are preserved in $\mathbf{z}^i,\forall i\in\{1,2,\cdots,L-1\}$ after passing through the ReLU layer. The two parts are then added together and split again in each layer until the weight $\mathbf{W}_{\text{O}}$ is applied in the final layer. Specifically, we set
\begin{align}
    &[\boldsymbol{\mu}_{\mathbf{w}^1_1},\boldsymbol{\mu}_{\mathbf{w}^1_2},\cdots,\boldsymbol{\mu}_{\mathbf{w}^1_{n_1}}]=\begin{bmatrix}
        \mathbf{I}_{d} & \boldsymbol{0}_{d\times(n_1-2d)} & -\mathbf{I}_{d}\\
        \boldsymbol{0}_{1\times d} & \boldsymbol{0}_{1\times (n_1-2d)} & \boldsymbol{0}_{1\times d}
    \end{bmatrix},\label{eq:initialization_1}\\
    &[\boldsymbol{\mu}_{\mathbf{w}^i_1},\boldsymbol{\mu}_{\mathbf{w}^i_2},\cdots,\boldsymbol{\mu}_{\mathbf{w}^i_{n_1}}]=\nonumber\\
    &\begin{bmatrix}
        \mathbf{I}_d\\
        \boldsymbol{0}_{(n_{i-1}-2d)\times d}\\
        -\mathbf{I}_d\\
        \boldsymbol{0}_{1\times d}
    \end{bmatrix}\begin{bmatrix}
        \mathbf{I}_{d} & \boldsymbol{0}_{d\times(n_i-2d)} & -\mathbf{I}_{d}
    \end{bmatrix},\nonumber\\
    &\forall i\in\{2,3,\cdots,L-1\},\label{eq:initialization_mid}\\
    &[\boldsymbol{\mu}_{\mathbf{w}^L_1},\boldsymbol{\mu}_{\mathbf{w}^L_2},\cdots,\boldsymbol{\mu}_{\mathbf{w}^L_{n_1}}]=\nonumber\\
    &\begin{bmatrix}
        \mathbf{I}_d & \boldsymbol{0}_{d\times 1}\\
        \boldsymbol{0}_{(n_{L-1}-2d)\times d} & \boldsymbol{0}_{(n_{L-1}-2d)\times 1}\\
        -\mathbf{I}_d & \boldsymbol{0}_{d\times 1}\\
        \boldsymbol{0}_{1\times d} & 1
    \end{bmatrix}\mathbf{W}_{\text{O}}.\label{eq:initialization_L}
\end{align}

\subsubsection{Weight Update}
We use the Rauch-Tung-Striebel (RTS) smoother to iteratively update the estimated distribution for $\mathbf{z}$, $\mathbf{u}$ and $\mathbf{w}$ for each layer. During the $k$-th update, the RTS equations at each layer $i\in\{L,L-1,\cdots,1\}$ are given by
\begin{align}
    \mathbf{K}_{\mathbf{u}^i}=&\boldsymbol{\Sigma}_{\mathbf{u}^i,\mathbf{z}^i}\boldsymbol{\Sigma}_{\mathbf{z}^i,\mathbf{z}^i}^{-1}\label{eq:backward_k_u}\\
    \boldsymbol{\mu}^+_{\mathbf{u}^i}=&\boldsymbol{\mu}_{\mathbf{u}^i}+\mathbf{K}_{\mathbf{u}^i}(\boldsymbol{\mu}^+_{\mathbf{z}^i}-\boldsymbol{\mu}_{\mathbf{z}^i})\label{eq:backward_mu_u}\\
    \boldsymbol{\Sigma}^+_{\mathbf{u}^i,\mathbf{u}^i}=&\boldsymbol{\Sigma}_{\mathbf{u}^i,\mathbf{u}^i}+\mathbf{K}_{\mathbf{u}^i}(\boldsymbol{\Sigma}^+_{\mathbf{z}^i,\mathbf{z}^i}-\boldsymbol{\Sigma}_{\mathbf{z}^i,\mathbf{z}^i})\mathbf{K}_{\mathbf{u}^i}^T\label{eq:backward_sigma_u}\\
    \mathbf{K}_{\mathbf{w}^i}=&\boldsymbol{\Sigma}_{\mathbf{w}^i,\mathbf{u}^i}\boldsymbol{\Sigma}_{\mathbf{u}^i,\mathbf{u}^i}^{-1}\label{eq:backward_k_w}\\
    \boldsymbol{\mu}^+_{\mathbf{w}^i}=&\boldsymbol{\mu}_{\mathbf{w}^i}+\mathbf{K}_{\mathbf{w}^i}(\boldsymbol{\mu}^+_{\mathbf{u}^i}-\boldsymbol{\mu}_{\mathbf{u}^i})\label{eq:backward_mu_w}\\
    \boldsymbol{\Sigma}^+_{\mathbf{w}^i,\mathbf{w}^i}=&\boldsymbol{\Sigma}_{\mathbf{w}^i,\mathbf{w}^i}+\mathbf{K}_{\mathbf{w}^i}(\boldsymbol{\Sigma}^+_{\mathbf{u}^i,\mathbf{u}^i}-\boldsymbol{\Sigma}_{\mathbf{u}^i,\mathbf{u}^i})\mathbf{K}_{\mathbf{w}^i}^T\label{eq:backward_sigma_w}\\
    \mathbf{K}_{\mathbf{z}^{i-1}}=&\boldsymbol{\Sigma}_{\mathbf{z}^{i-1},\mathbf{w}^i}\boldsymbol{\Sigma}_{\mathbf{w}^i,\mathbf{w}^i}^{-1},\label{eq:backward_k_z}\\
    \boldsymbol{\mu}^+_{\mathbf{z}^{i-1}}=&\boldsymbol{\mu}_{\mathbf{z}^{i-1}}+\mathbf{K}_{\mathbf{z}^{i-1}}(\boldsymbol{\mu}^+_{\mathbf{w}^i}-\boldsymbol{\mu}_{\mathbf{w}^i})\label{eq:backward_mu_z}\\
    \boldsymbol{\Sigma}^+_{\mathbf{z}^{i-1},\mathbf{z}^{i-1}}=&\boldsymbol{\Sigma}_{\mathbf{z}^{i-1},\mathbf{z}^{i-1}}\nonumber\\
    &+\mathbf{K}_{\mathbf{z}^{i-1}}(\boldsymbol{\Sigma}^+_{\mathbf{w}^i,\mathbf{w}^i}-\boldsymbol{\Sigma}_{\mathbf{w}^i,\mathbf{w}^i})\mathbf{K}_{\mathbf{z}^{i-1}}^T\label{eq:backward_sigma_z}
\end{align}
where the superscript $+$ denotes the updated parameter, \textit{e.g.,} we have $\boldsymbol{\mu}_{\mathbf{u}_i}=\mathbb{E}[\mathbf{u}^i|\mathcal{D}^{k-1}]$ and $\boldsymbol{\mu}_{\mathbf{u}_i}^+=\mathbb{E}[\mathbf{u}^i|\mathcal{D}^{k}]$. The cross covariance matrices in \eqref{eq:backward_k_w} and \eqref{eq:backward_k_z} give the cross covariance between two variables before and after linear transformations, which are given by
\begin{align}
    \mathbf{u}^i=\begin{bmatrix}
        (\mathbf{z}^{i-1}_1)^T & \boldsymbol{0} & \cdots & \boldsymbol{0}\\
        \boldsymbol{0} & (\mathbf{z}^{i-1}_1)^T & \cdots & \boldsymbol{0}\\
        \vdots & \vdots & \ddots &\vdots\\
        \boldsymbol{0} & \boldsymbol{0} & \cdots & (\mathbf{z}^{i-1}_1)^T\\
        (\mathbf{z}^{i-1}_2)^T & \boldsymbol{0} & \cdots & \boldsymbol{0}\\
        \boldsymbol{0} & (\mathbf{z}^{i-1}_2)^T & \cdots & \boldsymbol{0}\\
        \vdots & \vdots & \ddots &\vdots\\
        \boldsymbol{0} & \boldsymbol{0} & \cdots & (\mathbf{z}^{i-1}_2)^T\\
        \vdots & \vdots & \vdots & \vdots\\
        (\mathbf{z}^{i-1}_y)^T & \boldsymbol{0} & \cdots & \boldsymbol{0}\\
        \boldsymbol{0} & (\mathbf{z}^{i-1}_y)^T & \cdots & \boldsymbol{0}\\
        \vdots & \vdots & \ddots &\vdots\\
        \boldsymbol{0} & \boldsymbol{0} & \cdots & (\mathbf{z}^{i-1}_y)^T
    \end{bmatrix}\mathbf{w}^i
\end{align}
and \eqref{eq:linear_u_to_w}. Therefore, the cross-covariance can be easily calculated similar to \eqref{eq:sigma_u_forward}. Regarding the cross-covariance in \eqref{eq:backward_k_u}, for simplicity, we only keep the diagonal terms, which is given by~\cite{wagner_kbnn_2023}
\begin{align}
    \boldsymbol{\Sigma}_{\mathbf{u}^i,\mathbf{z}^i}[j,j]=&\left((\boldsymbol{\mu}_{\mathbf{u}^i}[j])^2+\boldsymbol{\Sigma}_{\mathbf{u}^i,\mathbf{u}^i}[j,j]\right)\phi\left(\frac{\boldsymbol{\mu}_{\mathbf{u}^i}[j]}{\sqrt{\boldsymbol{\Sigma}_{\mathbf{u}^i,\mathbf{u}^i}[j,j]}}\right)\nonumber\\
    &+\boldsymbol{\mu}_{\mathbf{u}^i}[j]\boldsymbol{\Sigma}_{\mathbf{u}^i,\mathbf{u}^i}[j,j]\mathcal{N}(0;\boldsymbol{\mu}_{\mathbf{u}^i}[j],\boldsymbol{\Sigma}_{\mathbf{u}^i,\mathbf{u}^i}[j,j])\nonumber\\
    &-\boldsymbol{\mu}_{\mathbf{u}^i}[j]\boldsymbol{\mu}_{\mathbf{z}^i}[j].
\end{align}

\subsection{Proposed Algorithm}
The proposed algorithm to fine-tune the transformer is given in Algorithm~\ref{alg:training}. In the algorithm, lines 1-8 initialize the prior distribution of the weights such that the original transformer behavior is reproduced. Lines 12-18 compute the forward pass for one data sample. Lines 23-30 compute the backward pass and updates the distributions of the weight for that sample.

\begin{algorithm}
\caption{Fine-tuning}
\label{alg:training}
\begin{algorithmic}[1]
\Require trained transformer $T=T_2\circ T_1$ with weights, training data $\mathcal{D}=\{\{\mathbf{X}^1,\mathbf{Y}^1\},\{\mathbf{X}^2,\mathbf{Y}^2\},\cdots,\{\mathbf{X}^N,\mathbf{Y}^N\}\}$, covariance $\boldsymbol{\Sigma}_{\text{data}}$ of the data
\State $\boldsymbol{\mu}_{\mathbf{w}^1}\gets\eqref{eq:initialization_1}$
\State $\boldsymbol{\Sigma}_{\mathbf{w}^1,\mathbf{w}^1}\gets\epsilon \mathbf{I}$
\For{$i$ in $\{2,\cdots,L-1\}$}
\State $\boldsymbol{\mu}_{\mathbf{w}^i}\gets\eqref{eq:initialization_mid}$
\State $\boldsymbol{\Sigma}_{\mathbf{w}^i,\mathbf{w}^i}\gets\epsilon \mathbf{I}$
\EndFor
\State $\boldsymbol{\mu}_{\mathbf{w}^L}\gets\eqref{eq:initialization_L}$
\State $\boldsymbol{\Sigma}_{\mathbf{w}^L,\mathbf{w}^L}\gets\epsilon \mathbf{I}$
\For{$i$ in $\{1,2,\cdots,N\}$}
\State $\{\hat{\mathbf{H}},\hat{\mathbf{Y}}\}\gets\Call{Preprocessing}{\{\mathbf{X}^i,\mathbf{Y}^i\}}$
\State $[\mathbf{h}_1,\mathbf{h}_2,\cdots]^T\gets\hat{\mathbf{H}}$
\State $\boldsymbol{\mu}_{\mathbf{z}^0}\gets [(\mathbf{h}_1)^T,(\mathbf{h}_2)^T,\cdots]^T$
\State $\boldsymbol{\Sigma}_{\mathbf{z}^0,\mathbf{z}^0}\gets \epsilon \mathbf{I}$
\For{$j$ in $\{1,2,\cdots,L-1\}$}
\State Compute $\boldsymbol{\mu}_{\mathbf{u}^j}$, $\boldsymbol{\Sigma}_{\mathbf{u}^j,\mathbf{u}^j}$, $\boldsymbol{\mu}_{\mathbf{z}^j}$, and $\boldsymbol{\Sigma}_{\mathbf{z}^j,\mathbf{z}^j}$ using \eqref{eq:mu_u_forward}, \eqref{eq:sigma_u_forward}, \eqref{eq:mu_z_forward}, and \eqref{eq:sigma_z_forward}
\EndFor
\State Compute $\boldsymbol{\mu}_{\mathbf{u}^L}$ and $\boldsymbol{\Sigma}_{\mathbf{u}^L,\mathbf{u}^L}$ using \eqref{eq:mu_u_forward} and \eqref{eq:sigma_u_forward}
\State Estimate $\boldsymbol{\mu}_{\mathbf{p}}$, $\boldsymbol{\Sigma}_{\mathbf{p},\mathbf{p}}$, and $\boldsymbol{\Sigma}_{\mathbf{u}^L,\mathbf{p}}$ using \eqref{eq:softmax_mu_forward}, \eqref{eq:softmax_sigma_forward}, and \eqref{eq:softmax_cross_covariance}
\State $\boldsymbol{\mu}_{\mathbf{z}^L}\gets \boldsymbol{\mu}_{\mathbf{p}}$
\State $\boldsymbol{\Sigma}_{\mathbf{z}^L,\mathbf{z}^L}\gets \boldsymbol{\Sigma}_{\mathbf{p},\mathbf{p}}$
\State $\boldsymbol{\Sigma}_{\mathbf{u}^L,\mathbf{z}^L}\gets \boldsymbol{\Sigma}_{\mathbf{u}^L,\mathbf{p}}$
\State $[\mathbf{y}_1,\mathbf{y}_2,\cdots]^T\gets\hat{\mathbf{Y}}$
\State $\boldsymbol{\mu}^{+}_{\mathbf{z}^L}\gets [(\mathbf{y}_1)^T,(\mathbf{y}_2)^T,\cdots]^T$
\State $\boldsymbol{\Sigma}^{+}_{\mathbf{z}^L,\mathbf{z}^L}\gets \boldsymbol{\Sigma}_{\text{data}}$
\For{$j$ in $\{L,L-1,\cdots,1\}$}
\State Compute $\boldsymbol{\mu}^{+}_{\mathbf{u}^j}$ and $\boldsymbol{\Sigma}^{+}_{\mathbf{u}^j,\mathbf{u}^j}$ using \eqref{eq:backward_mu_u}, \eqref{eq:backward_sigma_u}, and \eqref{eq:backward_k_u}
\State Compute $\boldsymbol{\mu}^{+}_{\mathbf{w}^j}$ and $\boldsymbol{\Sigma}^{+}_{\mathbf{w}^j,\mathbf{w}^j}$ using \eqref{eq:backward_mu_w}, \eqref{eq:backward_sigma_w}, and \eqref{eq:backward_k_w}
\State Compute $\boldsymbol{\mu}^{+}_{\mathbf{z}^{j-1}}$ and $\boldsymbol{\Sigma}^{+}_{\mathbf{z}^{j-1},\mathbf{z}^{j-1}}$ using \eqref{eq:backward_mu_z}, \eqref{eq:backward_sigma_z}, and \eqref{eq:backward_k_z}
\State $\boldsymbol{\mu}_{\mathbf{w}^j}\gets\boldsymbol{\mu}^{+}_{\mathbf{w}^j}$
\State $\boldsymbol{\Sigma}_{\mathbf{w}^j,\mathbf{w}^j}\gets \boldsymbol{\Sigma}^{+}_{\mathbf{w}^j,\mathbf{w}^j}$
\EndFor
\EndFor
\end{algorithmic}
\end{algorithm}

\section{Numerical Simulation}
\subsection{Settings}
We perform numerical simulation on a supervised learning problem on the decision transformer, where the input is the state of a dynamical system, and the output is the control action. In the fine-tuning process to adapt to changes in the dynamical system, the decision transformer, pre-trained with data from an optimal linear quadratic regulator (LQR) controller for the dynamical system before change, needs to adapt to the optimal LQR controller for the system after change. Specifically, we use an inverted pendulum system and evaluate the performance of the decision transformer using the success rate of stabilizing the system. The inverted pendulum system is characterized by 3 parameters: the mass of the cart $m_c$, the mass of the pendulum $m_p$, and the length from the cart to the pendulum's center of mass $l_p$. The states of the inverted pendulum are given by $\mathbf{x}_p:=[x_c,v_c,\theta_p,\omega_p]$, which are the location of the cart, the speed of the cart, the angle of the pendulum, and the angular velocity of the pendulum, respectively. The decision transformer has a 2-layer encoder with 2 attention heads, feedforward dimension of $8$, and a hidden layer dimension of $16$. The decision transformer takes as input the state and generates a control action. It is pre-trained with data from an optimal LQR controller for a linearized model of the inverted pendulum with parameters $m_c=1$, $m_p=0.1$, and $l_p=0.5$. During the fine-tuning process, the data are from an optimal LQR controller for a linearized model of the inverted pendulum with parameters $m_c=1$, $m_p=1$, and $l_p=5$, and the fine-tuned decision transformer is expected to be able to stabilize a system with such parameters. Note that we consider the data available for fine-tune to come from i.i.d. samples of the optimal controller, instead of trajectories from system controlled by the optimal controller, which do not give i.i.d. data. We compare the proposed method with warm-started retrainings of the transformer. Specifically, for the warm-started retraining, the model starts training with the parameters in the pre-trained model. To test the performance of the sequential learning process with limited memory, we test the memory capacities of $10$, $20$, $25$, $50$, $75$, and $100$ training samples. The model holds memory of one of the listed capacities, and trains for $100$ epochs with the samples in the memory. When new samples arrive, the old samples saved in the memory are removed. Note that the proposed method only requires a memory capacity of $1$ training sample. We perform $10$ trials for each method, where in each trial, each method receives a total of $400$ samples. In addition, to test the performance of the proposed method in uncertainty quantification, we use stochastic training data of different covariances $\boldsymbol{\Sigma}_{\text{data}}$ and record the predicted covariances for different training iterations (also number of samples processed). We test the following values for $\boldsymbol{\Sigma}_{\text{data}}$: $0$, $10$, $20$, and $50$.

\subsection{Results and Analyses}
\begin{figure}
    \centering
    \includegraphics[width=\linewidth]{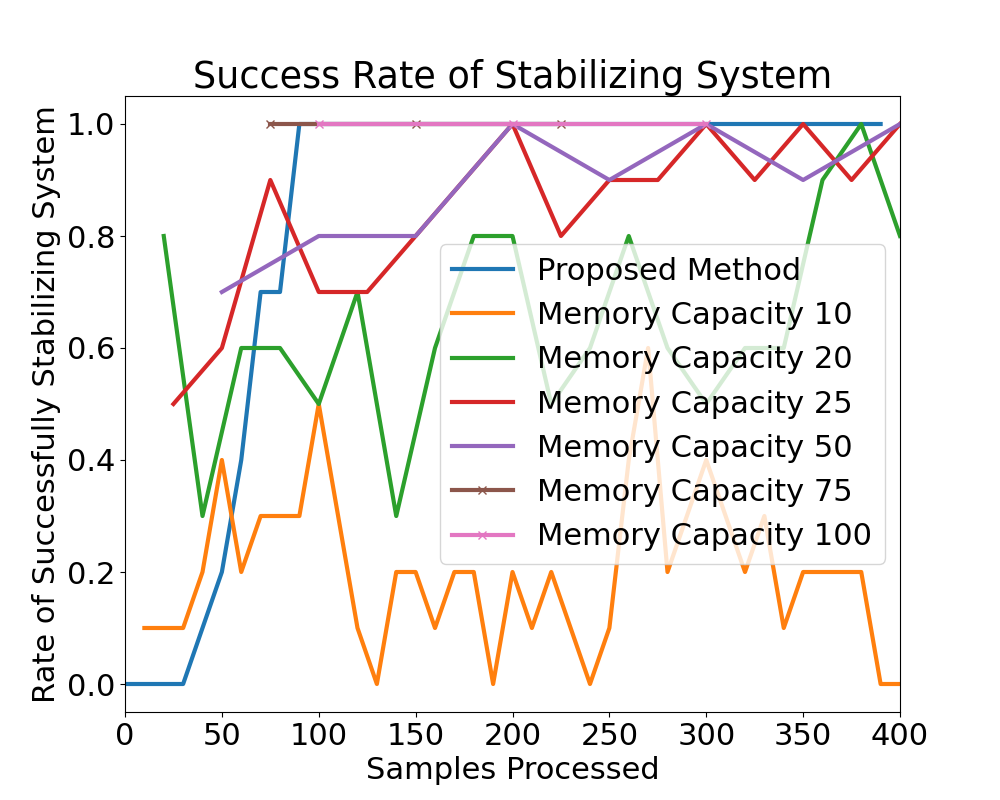}
    \caption{The rate of success in stabilizing the inverted pendulum. The blue line shows the proposed method, and the other lines show the warm-started retraining with different memory capacities.}
    \label{fig:success_rate}
\end{figure}

\begin{figure}
    \centering
    \includegraphics[width=\linewidth]{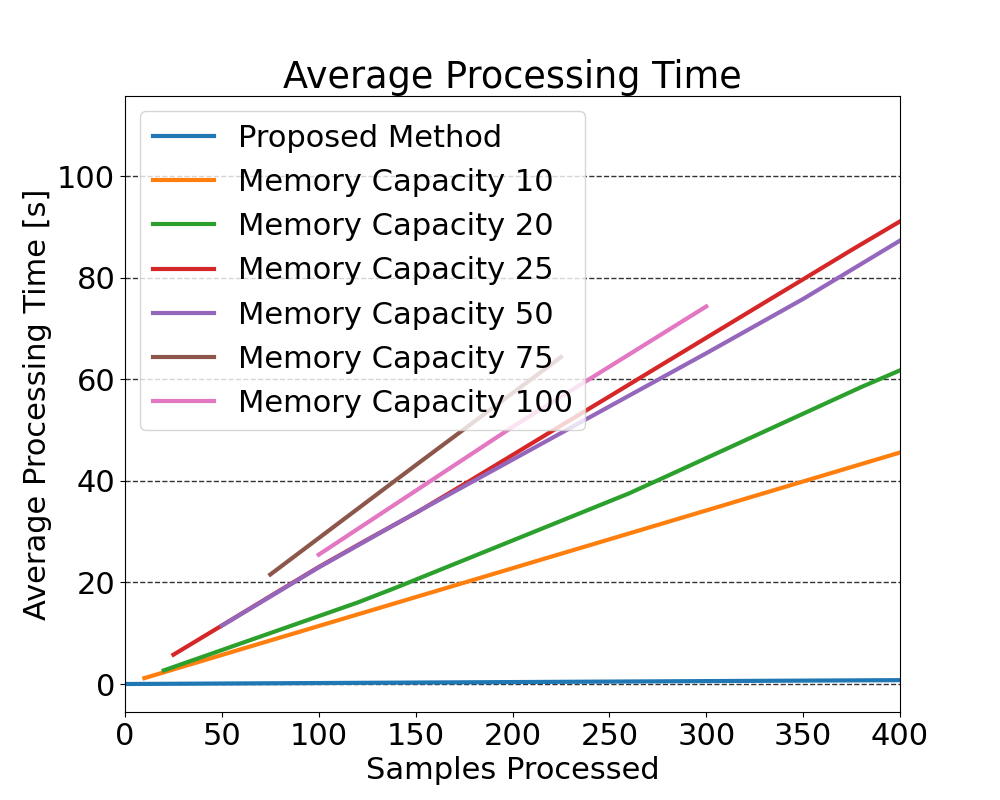}
    \caption{The average computation time. The blue line shows the proposed method, and the other lines show the warm-started retraining with different memory capacities.}
    \label{fig:computation_time}
\end{figure}

\begin{figure}
    \centering
    \includegraphics[width=\linewidth]{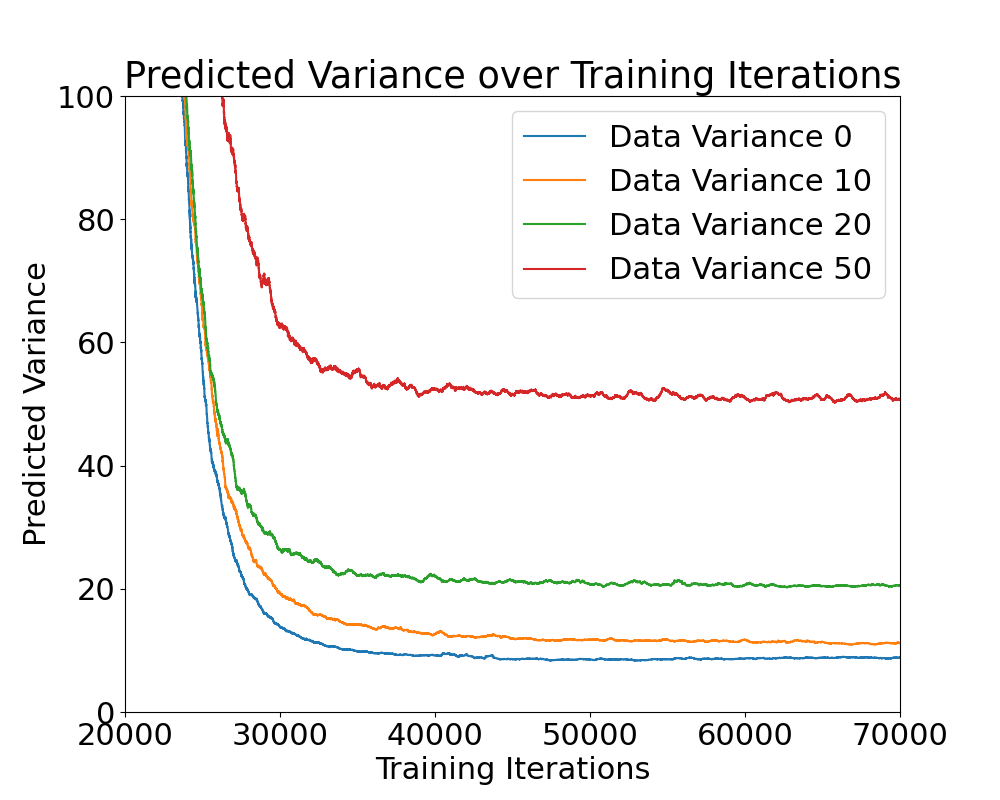}
    \caption{The uncertainty predicted by the proposed method versus training iterations (number of samples processed). To ensure a clear view of the data in the presence of uncertainties, we apply a moving average filter of size $500$ and plot the training iterations $30000$ to $70000$, during which the variances stop to decrease.}
    \label{fig:uncertainty_quantification}
\end{figure}

We first record the success rate of stabilizing the inverted pendulum system for both the memory-constrained warm-started retraining method and the (more memory constrained) proposed method, for different numbers of training samples. The results are shown in Figure~\ref{fig:success_rate}. While only requiring a memory capacity of $1$ sample, the proposed method outperforms the warm-started retraining method with memory capacities $10$, $20$, $25$, and $50$ in terms of success rate. Note that the retraining method frequently sees drops in success rate as more data arrive. This is due to the catastrophic forgetting phenomenon of the neural network. On the other hand, the proposed method does not experience drops, due to Bayesian neural network's capability of preventing catastrophic forgetting~\cite{li2020continual}. We also record the training time versus number of training samples for both methods (Figure~\ref{fig:computation_time}). The proposed method has a significantly lower computation time per sample compared to the retraining method with all memory capacities. Figure~\ref{fig:uncertainty_quantification} shows the result for uncertainty quantification. The predicted uncertainty increases as the data uncertainty increases, and the predicted uncertainty matches the data uncertainty better at higher uncertainties. This is due to that there exists a nonzero prediction uncertainty for data points that are not equal to any data points in the training dataset (akin to uncertainties in Gaussian process regression). When the data uncertainty is large, this uncertainty is dominated by the data uncertainty.

\section{Conclusion and Future Work}
In this paper, we propose a method that formulates sequential fine-tuning problem for transformers as a Bayesian posterior inference problem. The proposed method integrates closed-form moment propagation of random variables, Kalman Bayesian Neural Networks, and Taylor approximations of the moments of softmax functions. In this way, the proposed method has the advantage of memory-efficient and computation-efficient sequential learning with explicit characterization of the uncertainties in the prediction. We demonstrate the effectiveness of the proposed method using numerical simulation.
\subsection{Limitations and Future Work}
One of the major limitations of the proposed method is regarding the off-diagonal terms in the covariance propagation. While we include the off-diagonal terms in the covariance matrix in the forward pass, there still does not exist a closed-form propagation method for the off-diagonal terms in the backward pass through the nonlinear activation layers. Accounting for such terms would be a valuable future direction.

\bibliographystyle{ieeetr}
\bibliography{bibliography}

\end{document}